\documentclass[journal]{IEEEtran}

\usepackage{color}
\usepackage{graphicx}
\usepackage{amssymb}
\usepackage{epstopdf}
\usepackage{amsmath}
\usepackage{subcaption}
\usepackage{bm}
\usepackage{amsmath}
\usepackage[lined,boxed,linesnumbered,noend]{algorithm2e}
\usepackage{multirow}
\usepackage{cite}
\usepackage{verbatim}

\usepackage{dcolumn}
\newcolumntype{N}{@{}m{0pt}@{}}

\usepackage{array}

\usepackage{bigstrut}
\newcolumntype{C}[1]{>{\centering\let\newline\\\arraybackslash\hspace{0pt}}m{#1}}
\newcommand{\specialcell}[2][c]{%
  \begin{tabular}[#1]{@{\bigstrut[t]}c@{}}#2\end{tabular}}

\title{\LARGE \bf Slip Detection: Analysis and Calibration of Univariate Tactile Signals }

\author{Karl~Van~Wyk,
        Joe~Falco% <-this % stops a space
\thanks{}% <-this % stops a space
\thanks{K. Van Wyk and J. Falco are with the National Institute of Standards and Technology, Gaithersburg, MD, USA (e-mail: karl.vanwyk@nist.gov; joseph.falco@nist.gov). \newline
\indent \textit{Disclaimer: }Certain commercial equipment, instruments, or materials are identified in this 
paper to foster understanding. Such identification does not imply recommendation 
or endorsement by the National Institute of Standards and Technology, nor does it imply that the materials or equipment identified are necessarily the best available for the purpose.}}

\begin{document}

\maketitle
\thispagestyle{empty}
\pagestyle{empty}

% write the abstract with the Abstract-environment:

\begin{abstract}
%Background
The existence of tactile afferents sensitive to slip-related mechanical transients in the human hand augments the robustness of grasping through secondary force modulation protocols. Despite this knowledge and the fact that tactile-based slip detection has been researched for decades, \textit{robust} slip detection is still not an out-of-the-box capability for any commercially available tactile sensor. This research seeks to bridge this gap with a comprehensive study addressing several aspects of slip detection. Key developments include a systematic data collection process yielding millions of sensory data points, the generalized conversion of multivariate-to-univariate sensor output, an insightful spectral analysis of the univariate sensor outputs, and the application of Long Short-Term Memory (LSTM) neural networks on the univariate signals to produce robust slip detectors from three commercially available sensors capable of tactile sensing. The sensing elements underlying these sensors vary in quantity, spatial arrangement, and mechanics, leveraging principles in electro-mechanical resistance, optics, and hydro-acoustics. Critically, slip detection performance of the tactile technologies is quantified through a measurement methodology that unveils the effects of data window size, sampling rate, material type, slip speed, and sensor manufacturing variability. Results indicate that the investigated commercial tactile sensors are inherently capable of high-quality slip detection.

\end{abstract}


\IEEEpeerreviewmaketitle

\section{Introduction}
%Motivation
\IEEEPARstart{N}{europhysiological} research reveals the existence of four distinct types of tactile afferents in the human hand: fast-adapting type I (FA-I), slow-adapting type I (SA-I), fast-adapting type II (FA-II), and slow-adapting type II (SA-II) \cite{johansson2009coding}. This variety of tactile afferents affords a spectrum of sensitivity to various mechanical stimuli, an arrangement determined critical for proper sensorimotor control of the hand. Among the many functional modalities these afferents impart, one of particular interest is the sensing of high-frequency mechanical transients (5 Hz - 400 Hz) via FA-I and FA-II type afferents. For example, these vibrations can emanate from surfaces in sliding contact that undergo quick transitions among friction states, also known as the ``catch and snap'' effect \cite{howe1989sensing}. Naturally, humans can detect and react to object slip through secondary force modulation efforts with a reflex time of less than 100 msec \cite{Johansson1984}. In essence, this sensorimotor response acts as a `fail-safe' in the event that primary force modulation protocols do not suffice. Imbuing robotic hands and grippers with tactile-enabled slip detection behaviors is one logical avenue for elevating the robustness of robotic grasping.

%Review of existing sensors
%Review of studies for calibration of slip
Tactile slip detection is not a novel concept; investigative efforts started in the late 1980s and continue to the modern day. To date, researchers have produced slip detectors from wildly different sensors including accelerometers \cite{howe1989sensing}, force transducers \cite{yamada1994tactile}, pressure-sensitive tactile arrays \cite{holweg1996slip,alcazar2012estimating}, piezoelectric polymer films \cite{goger2009tactile,heyneman2016slip}, elastomer-embedded cameras \cite{yuan2015measurement}, carbon nanotube-polymer composites \cite{vatani2013force}, pressure transducers \cite{su2015force}, capacitance arrays \cite{heyneman2016slip}, and strain gauges \cite{fernandez2014micro}. The majority of slip detector algorithms are at least partially composed of spectral analyses from Fourier and wavelet transforms, to extract relevant features for slip classification \cite{holweg1996slip,fernandez2014micro,heyneman2016slip}. Other approaches apply optical flow algorithms \cite{alcazar2012estimating}, band-pass filters \cite{su2015force}, or contact force cone and force measurements to predict slip \cite{kaboli2016tactile}. Reported slip classification accuracies were above 90 \% in \cite{su2015force} and above 85 \% in \cite{heyneman2016slip}.

Despite such a rich variety of sensor designs and algorithmic approaches for slip detection, commercial tactile sensors are still devoid of this capability. The resulting implication is that slip-detecting tactile sensors have not yet approached readiness levels for commercialization. This result is likely due to both methodological and algorithmic inadequacies. Methodologically, slip detection research often does not fully investigate detection performance across the major factors (window size, sampling rate, slip speed, contacting materials, latent vibrations, contact force, manufacturing variability), a trend noticed by others as well \cite{agriomallos2018slippage}. Algorithmically, subsequent slip detection accuracies are purely insufficient for applied controls. The former issue can be resolved by a thorough design of experiments across all relevant factors in order to systematically hone and test slip detector quality. The latter issue can be addressed with the application of more powerful algorithms capable of analyzing sequences of sensor data for classifying slip events with a high degree of accuracy.

%Difficult problem, advances in AI, namely, deep nets - powerful algorithms for analyze temporal sequences of data
In the era of high performing neural networks in challenging problem domains, strategies for producing highly robust slip detection for tactile sensors should include neural networks and automated data acquisition of large datasets. Coupling deep neural networks and extensive datasets with contemporary computational power have led to significant performance gains in computer vision \cite{krizhevsky2012imagenet}, language modeling \cite{melis2017state}, language translation \cite{wu2016google}, artificial gaming agents \cite{mnih2015human}, bin picking \cite{levine2016learning}, and speech synthesis \cite{oord2016wavenet}. Since slip detection is primarily imparted through temporal signatures of tactile data \cite{Johansson1984,agriomallos2018slippage,veiga2015stabilizing}, the most promising neural architecture for slip detection is \textit{recurrent} neural networks. 

By design, recurrent neural networks excel at processing \textit{sequences} of data through the existence of internal neuronal states in the hidden layers that integrate input data over the entire input sequence. Recurrent networks have been notoriously difficult to train in the past due to their exploding or vanishing gradients. However, this issue has been largely resolved by Long Short-Term Memory (LSTM) recurrent networks \cite{greff2017lstm}. LSTM networks circumvent these numerical issues during the training process through their intrinsic dynamics, and excel at finding patterns in data separated by large time intervals - an attractive trait for slip detection where several hundred sensor readings may be necessary for maximizing accuracy.

%Paper contributions and outline
This research is the first to apply LSTM networks to large datasets gathered over the previously listed factors to create robust slip detectors. The main contributions include: 1) a systematic design of experiments for collecting millions of data points for training slip detectors; 2) an in-depth, comparative spectral analysis of each sensor, yielding insight on the most significant frequencies for slip detection; 3) conversion to univariate tactile signals and application of LSTM networks for slip detection; and 4) a measurement methodology that analyzes windows size, sampling rate, material, slip speed, and manufacturing variability effects on slip detection accuracy.

%Section discussing hardware

\section{Experimentation}
\label{sec:data_collection}

\subsection{Hardware}
The hardware components (commercial products) included the KUKA Light Weight Robot 4+; the Wonik Allegro robotic hand; and the Nano17, OptoForce20, and Biotac SP tactile sensors. The arm was Cartesian position controlled, and the hand was joint impedance controlled (proportional-derivative controller with gravity model) with a joint stiffness of 3 Nm/rad, joint damping of 0.075 Nm$\cdot$s/rad, and torque saturation limits of $\pm 0.42$ Nm.  Each set of tactile sensors was mounted as the fingertips (i.e., index, middle, little) of the robotic hand attached to the arm as depicted in Fig. \ref{fig:Collage}. Nano17s (three-count) were six-axis, silicon strain gauge-based force-torque transducers with a 3D printed, rubber-coated fingertip and a force sensitivity of 6 mN. OptoForce20s OMD-20-SE-40N (three-count) were three-axis force sensors that used optics to measure mechanical deformation of the outer rubber dome with a force sensitivity of 2.5 mN. Biotac SPs (two-count) were biomimetic sensors with a fluid-filled rubber membrane encapsulating a rigid core that housed multiple electrodes, a pressure transducer, and a thermistor with a force sensitivity of at least 10 mN.  Data was collected from the six strain gauge signals at 1000 Hz for the Nano17s, the three optic signals at 1000 Hz for the OptoForce20s, and only the high-pass filtered pressure transducer signal at 850 Hz for the Biotac SPs (designed for high-frequency signal applications).

\subsection{Factors}
To train robust slip detectors with LSTM networks, a large, highly representative dataset was necessary. Specifically, relevant factors that affect a slip detector's performance in both slip and non-slip events must be known. \textit{Controllable} factors for this study included sensor type and manufacturing variability, sampling rate, window size, extrinsic material surfaces, slip speed between sensors and extrinsic surfaces, ambient or motion vibration from the sensor's connecting structures, and the sensor's force-loading profile. 

\subsection{Data Collection}
The data collection process was designed to concurrently sample tactile data across all instances of the same sensor type for a variety of non-slipping and slipping scenarios. The experiments were constrained to collect data during interactions with the primary contact areas or ``center'' of the tactile sensors, away from their outer perimeter connections where sensitivity is often diminished. Primary areas are the most likely to make contact with extrinsic surfaces due to their location and typically convex shape. 

%Way to create two-column picture
\begin{figure}
\centering
\includegraphics[width=1\linewidth]{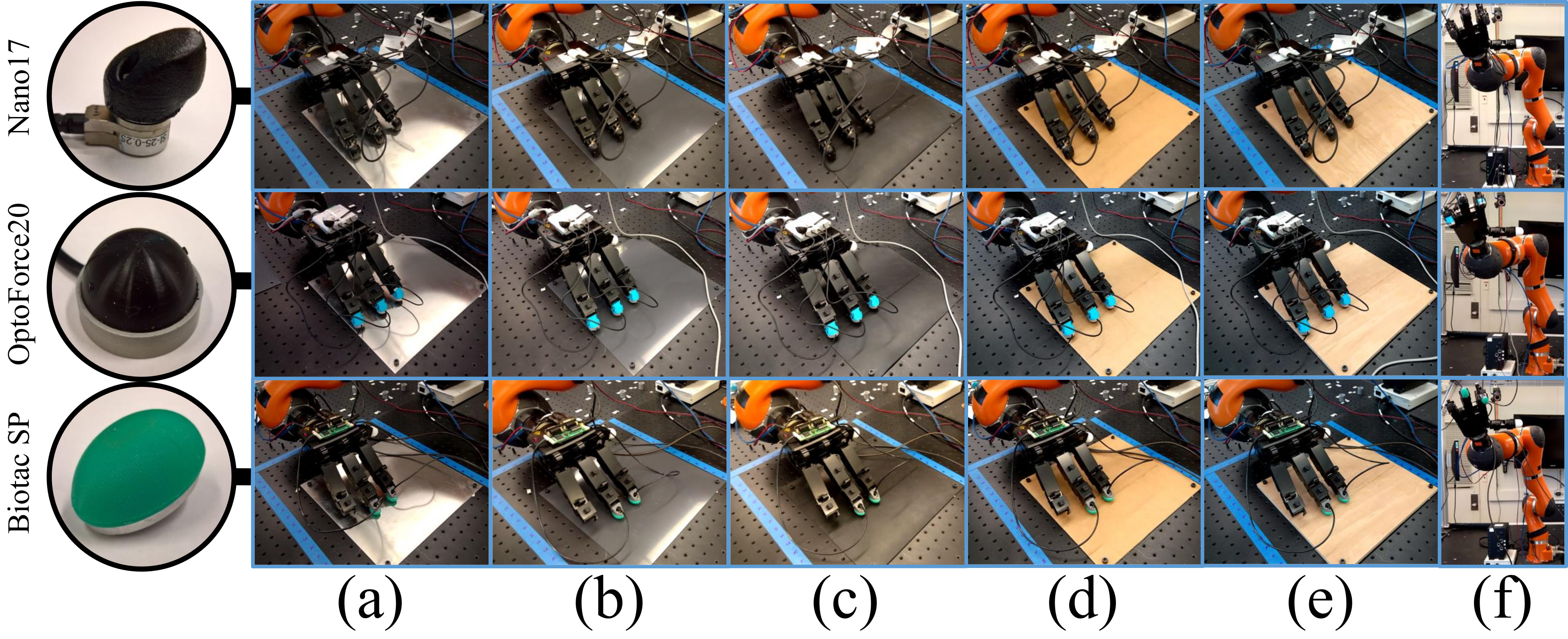}
\caption{Gathering tactile data on (a) aluminum, (b) PVC, (c) neoprene, (d) cardboard, (e) plywood, and in (f) free-space.}
\label{fig:Collage}
\end{figure}

\subsubsection{Non-Slipping Data}
Two scenarios were engaged to collect non-slipping tactile data. The first scenario involved repositioning the arm and hand to 36 different joint configurations in free space (see Fig. \ref{fig:Collage}f) at three different speed settings, yielding \textit{average} end-of-arm speeds of 25 mm/s, 50 mm/s, and 75 mm/s. Each speed setting was sequentially applied to the same set of 36 joint configurations. Finger joint configurations were generated via the Latin Hypercube sampler with joint velocities pseudo-randomly sampled between 2 deg/s and 10 deg/s. The data generated for this scenario were labeled as non-slipping data since the tactile sensors did not make surface contact. This data helped reduce false-positive slip detection rates originating from the idling vibration or motion of the connecting electro-mechanical structures (i.e., arm and hand).

Another case of non-slipping data included force modulating the sensors at varying frequencies and magnitudes with a prospective surface in order to emulate scenarios in which the robotic hand or gripper readjusted its grasp or performed in-hand manipulation on an object without slipping. The robotic arm statically positioned the robotic hand within reach of five different material sheets: aluminum, polyvinyl chloride (PVC), neoprene, cardboard, and plywood. The fingers were commanded to initiate contact and palpate the materials at various force levels and frequencies by commanding the fingers into the working surface at 100 different inwardly-curled joint angle configurations and speeds sampled from the Latin Hypercube sampler. Joint angle changes of 0 deg to 70 deg were issued across flexing joints of the contacting fingers. Joint speeds ranged from 0.125 deg/s to 25 deg/s.

\subsubsection{Slipping Data}
To capture the tactile sensory response to slipping events, variations in slip speed, material, and contacting force must be represented. Accordingly, the robotic arm was commanded to drag the hand back-and-forth along a linear path (18 cm in length) with tactile-surface contact across various material types 32 times at four speed settings, yielding \textit{average} end-of-arm speeds of 5 mm/s, 25 mm/s, 50 mm/s, and 75 mm/s. This purely lateral motion preventing the occurrence of finger jamming into the surfaces, and mimicked sensor-material motions of other slip studies \cite{kaboli2016tactile,agriomallos2018slippage,fernandez2014micro,heyneman2016slip}. Although only four arm speed settings were used, the arm path generator creates motion paths that smoothly changes the arm speed between zero and a preset maximum, yielding the aforementioned average speeds and a continuous spectrum of speed over the planned path. Joint angles to the hand were held constant per set of tactile sensors that yielded appreciable contact with the materials. Deviations in hand joint angles were not issued to: 1) prevent accidental fingertip-surface lift; 2) reduce excessive wear on the sensor surfaces; and 3) prevent mechanical damage to the fingers from large contact forces and lateral arm accelerations. However, contact force and angle were passively varied by the natural oscillation of the moving arm, and the twisting and rolling of the compliant fingers on the materials from the sliding cycles.

\subsection{Data Balancing}
\label{subsec:data_balancing}
After acquisition, the data was balanced to prevent a classifier from exhibiting biases towards more frequently represented patterns and split into training and testing datasets.  Overall, a representative dataset consisted of a 50-50 \% split of non-slipping and slipping data. With five materials and four speed settings, each material yielded 20 \% of the slipping data, and each slip speed setting yielded 25 \% of each material. Moreover, pushing data and free-space data constituted 50 \% of non-slipping data, each. Each material yielded 20 \% of the pushing data. Each speed setting yielded 33 \% of free-space data. To enforce these proportions, the original datasets were trimmed from the end, shuffled based on a pre-selected window size (i.e., the number of sequential sample readings), concatenated based on type (e.g., slipping or non-slipping), and shuffled again based on a pre-selected window size. This yielded about eight million data points (sensor samples) for every individual sensor -- half for training and half for testing.

\subsection{Data Preprocessing}
A unified approach was taken to produce a single, high-frequency data stream from a sensor's raw tactile data. Since a general force-sensitive tactile sensor may have multiple data outputs, a formula for obtaining a single data stream from all relevant sensor outputs involves calculating the lag-one finite difference of the \textit{l}$_2$-norm of the sensor vector data,

\begin{equation} 
\label{eq:l2_norm}
s_t = ||\mathbf{a}_t||_2-||\mathbf{a}_{t-1}||_2 \quad
\end{equation}

\noindent where $s_t \in \mathbb{R}$ is the preprocessed sensor signal at timestep $t$; $\mathbf{a}_t,\mathbf{a}_{t-1} \in \mathbb{R}^{m}$ are the m-dimensional sensor vector data at timestep $t$ and $t-1$, respectively; and $||\cdot||_2$ denotes the \textit{l}$_2$-norm. This formulation effectively pools the full magnitude of the sensor's response to \textit{changes} in force during tactile experiences, a signal gradient inspired by the fast-acting tactile afferents in the human hand. Furthermore, this collapsed, lag-one difference produces a data stream centered around zero that signals changes in force loading while removing perception of absolute force magnitudes and direction. Effectively, this filters out unintentional biases from the data collection process that can lead to training slip detectors from undesirable patterns. For instance, slip detectors trained from the raw 3D data of the Nano17s and OptoForce20s could have learned to detect slip purely from the magnitudes or direction of the lateral forces, neither of which necessarily mandate the existence of slip. Instead, using a univariate signal gradient forces slip detectors of the Nano17s and OptoForce20s to learn purely from collapsed signal changes like that of the Biotac SPs, where only a single high-pass filtered pressure signal is utilized without context of force magnitude and direction.

\section{Spectral Analysis}
\label{sec:spectral_analysis}
While LSTM networks are used to perform slip detection, an in-depth spectral analysis of the univariate sensor data was first conducted to characterize sensory response and provide sensor behavior insights. Quantifying frequency bands that yield statistically significant differences in sensory response between non-slipping and slipping data can indicate the feasibility for slip calibration and guide sensor sampling rates.

\begin{figure*}[ht]
\centering
\includegraphics[width=.95\textwidth]{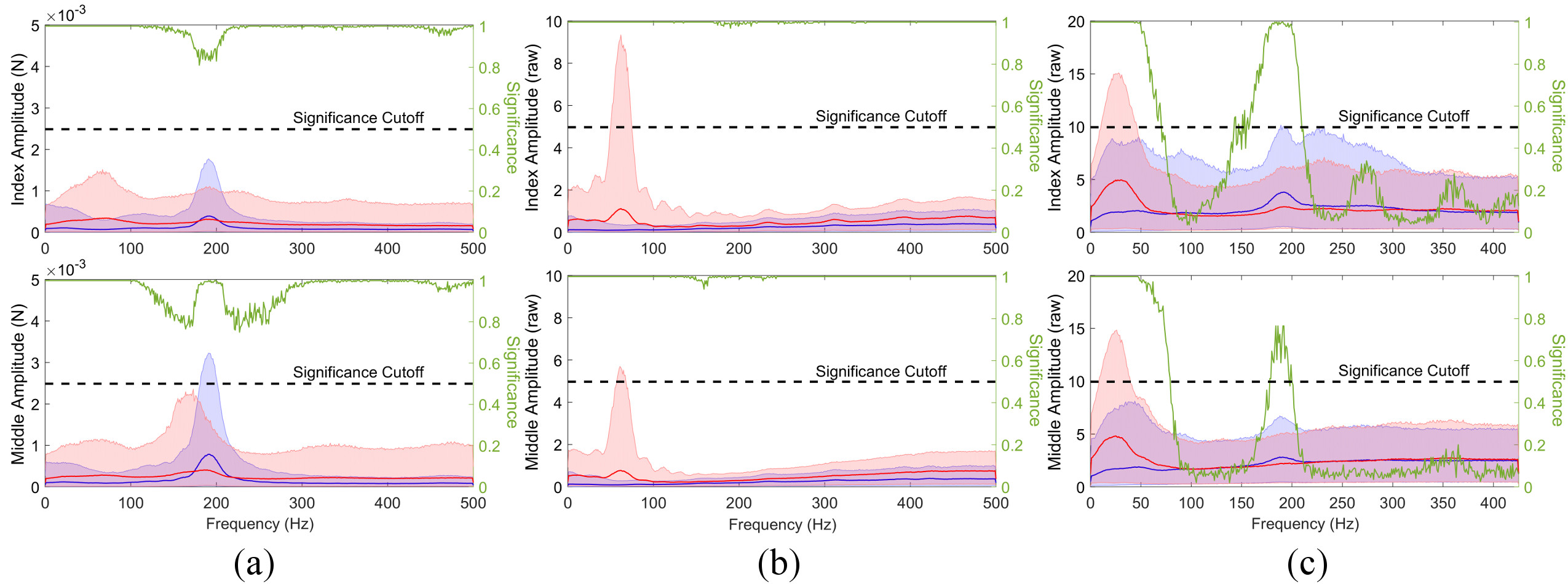}
\caption{Spectral plots of index and middle (a) Nano17, (b) OptoForce20, and (c) Biotac SP indicating the mean amplitude (solid lines), 95 \% confidence bands (faded areas), and significance (green line). Blue is non-slipping data, red is slipping data.}
\label{fig:spectral_analysis}
\end{figure*}

After data balancing and preprocessing, the collected data for each sensor was pooled into one of two categories: non-slipping and slipping.  Next, both non-slipping and slipping datasets were bootstrap-sampled 100 times, wherein each sample consisted of a sequence of sensor data of equal length to the maximum sampling rate of the sensor (e.g., 1000 data points per sequence for the Nano17s and OptoForce20s). Each non-slipping and slipping sequence was analyzed by the Fast Fourier Transform, yielding a single-sided amplitude spectrum at 1 Hz resolution over a frequency range of zero to one-half the maximum sensor sampling rate. The returned amplitudes per frequency for each set of bootstrapped samples were analyzed by the two-sample Kolmogorov-Smirnov (KS) algorithm between non-slipping and slipping data. The KS algorithm is a statistical, non-parametric distribution test for continuous data that determines whether or not two sample sets belong to the same population \cite{stephens1974edf}. By comparing 100 samples of non-slipping and slipping amplitudes per frequency, the KS algorithm exposed those frequencies for which the sensor data was the most distinguished between non-slipping and slipping events. This bootstrap sampling process and statistical test was repeated 200 times, yielding 200 KS tests per frequency. A single KS test either accepted or rejected the null hypothesis by returning a value of 0 or 1, respectively. Averaging the 200 KS test verdicts per frequency yielded a stable \textit{significance} signal as shown in Fig. \ref{fig:spectral_analysis}. Also shown are the mean and 95 \% confidence bands of the amplitudes, indicating the variability in sensory response during non-slipping and slipping events.

A number of important results are evident in Fig. \ref{fig:spectral_analysis}. Every sensor type exhibited contrasting features for non-slipping and slipping data. The Nano17s and OptoForce20s yielded significant differences in signal amplitudes across the entire range of frequencies (0 Hz - 500 Hz). In contrast, the Biotac SPs expressed two significant bands of frequencies including 0 Hz through 75 Hz and 175 Hz through 225 Hz. These frequencies fall directly within the sensing realm of FA-I afferents (5 Hz - 50 Hz) and FA-II afferents (40 Hz - 400 Hz). At approximately 200 Hz, both the Nano17s and Biotac SPs experienced spikes in amplitudes for non-slipping data, the known idling vibration of the robotic arm. The OptoForce20s appeared completely insensitive to the arm's idling vibration (likely due to the mechanical isolation and compliance of the sensor's air-filled rubber dome with only perimeter connections to the sensor base). Additionally, the most significant frequency band ($\sim$1 significance) that generally contains the largest separation in amplitude means is 0 Hz - 100 Hz for the Nano17s and OptoForce20s, and 0 Hz - 50 Hz for the Biotac SPs. Generally, both the Nano17s and OptoForce20s experienced peak separation in mean curves at approximately 60-70 Hz, the dominant frequency of the catch-and-snap effect \cite{holweg1996slip}. Unlike the Nano17s, the slip state bands for both the OptoForce20s and Biotac SPs experienced increasing overlap beyond approximately 250 Hz. This effect was likely due to their thicker outer contacting rubber surfaces acting as mechanical low-pass filters to these very high frequency signals. The Nano17s did not exhibit this behavior likely due to its higher overall rigidity and a rubber coating less than 0.2 mm.

Overall, these plots provided strong evidence that simply setting amplitude thresholds from FFT analyses of sensor data could result in a brittle slip detection strategy since major overlaps in amplitudes existed across the full frequency range between non-slipping and slipping data for all sensor types. Moreover, large data collection is still necessary to create these high-fidelity spectral plots to guide a thresholding technique. Regardless, such a technique was briefly investigated in Section \ref{subsec:window_size} which confirmed a major loss in slip classification accuracy when compared with that of LSTM networks.

\section{Calibration for Slip Detection}
\label{sec:calibration}

Many existing approaches seek to extract relevant features from spectral algorithms that are either used directly to predict slip or to train a separate classifier to predict slip. Although these approaches have produced worthwhile results in the past, the feature extraction process may have failed to produce all relevant features in the data. Instead, this new approach passes minimally processed sensor data streams directly to the LSTM networks such that they are less likely to exclude obscured, yet relevant information for training a robust slip detector.

\subsection{LSTM Architecture}
The governing equations for the binary classification LSTM model (refer to \cite{greff2017lstm} for more details regarding LSTM cells) used for slip detection are: 

\begin{equation} 
\label{eq:LSTM_blockinput}
z^t = g(W_z x^t + R_z y^{t-1} + b_z)
\end{equation}
\begin{equation} 
\label{eq:LSTM_inputgate}
i^t = \sigma(W_i x^t + R_i y^{t-1} + b_i)
\end{equation}
\begin{equation} 
\label{eq:LSTM_forgetgate}
f^t = \sigma(W_f x^t + R_f y^{t-1} + b_f)
\end{equation}
\begin{equation} 
\label{eq:LSTM_cell}
c^t = z^t \odot i^t + c^{t-1} \odot f^t
\end{equation}
\begin{equation} 
\label{eq:LSTM_outputgate}
o^t = \sigma(W_o x^t + R_o y^{t-1} + b_o)
\end{equation}
\begin{equation} 
\label{eq:LSTM_blockoutput}
y^t = h(c^t) \odot o^t
\end{equation}
\begin{equation} 
\label{eq:net_out}
y_p^t = \sigma (W_y y^t + b_y),
\end{equation}

\noindent where $W_z$, $W_i$, $W_f$, $W_o \in \mathbb{R}^{N \times M}$ are the inputs weights; $R_z$, $R_i$, $R_f$, $R_o \in \mathbb{R}^{N \times N}$ are the recurrent weights; $b_z$, $b_i$, $b_f$, $b_o \in \mathbb{R}^{N \times 1}$ are the hidden bias weights; $W_y \in \mathbb{R}^{K \times N}$ is the output weights; and $b_y \in \mathbb{R}^{K \times 1}$ is the output bias weights. Furthermore, $x^t \in \mathbb{R}^{M \times 1}$ is the input vector; $z^t \in \mathbb{R}^{N \times 1}$ is the block input; $i^t \in \mathbb{R}^{N \times 1}$ is the input gate; $f^t \in \mathbb{R}^{N \times 1}$ is the forget gate; $c^t \in \mathbb{R}^{N \times 1}$ is the cell state; $o^t \in \mathbb{R}^{N \times 1}$ is the output gate; $y^t \in \mathbb{R}^{N \times 1}$ is the block output; and $y_p^t \in \mathbb{R}^{K \times 1}$ is the network output. Finally, $\sigma(\cdot)$ is the logistic sigmoid function; both $g(\cdot)$ and $h(\cdot)$ are the hyperbolic tangent function; and $\odot$ is the element-wise vector product. For binary classification, $K=2$, and each output is a number between 0 and 1, which is interpreted as the likelihood of belonging to that output's class (i.e., either non-slipping or slipping). For a single, preprocessed sensory signal, $M=1$. The number of hidden LSTM units, $N$, is determined below.

\subsection{LSTM Network Parameters and Optimization}
Through trial and error, the network parameters were fixed at values that produced relatively well-functioning slip detectors across all investigated sensors on the training datasets. The hidden layer size was fixed at 20 LSTM neurons, which yielded useful classification rates while minimizing training time and preventing overfitting. Momentum was held constant at 0.125. Mimicking a cooling schedule, four learning rates of 0.01, 0.001, 0.0001, and 0.00001 were applied sequentially for up to 100 training epochs, or until no measurable progress was demonstrated for 10 consecutive epochs. Optimization was conducted via full back-propagation through time (BPTT) by calculating the network gradients of a fully unfolded network. The typical cross-entropy loss function was used for optimization. The input data stream to the networks were normalized by the 2.5\textsuperscript{th} and 97.5\textsuperscript{th} percentiles of the training datasets rather than the minimum and maximum values to reduce detrimental squashing of input data from the extreme outliers in the data. Training and testing models were conducted on a commodity laptop with a CPU core speed of 2.8 GHz. Training time took less than 23 $\mu$s per sensor sample, e.g.,  about 100 s per epoch for a dataset of 4.4 million sensor samples. Training time was invariant of model window size since scaling window size reciprocally scaled the number of unique sensor sequences from a fixed dataset (see Section \ref{sec:sensor_paramters} on window size). Inference took less than 5 $\mu$s per sensor sample, e.g., less than 0.3 ms for a window size of 50 samples.

Since this study investigated slip detection performance over a variety of factors and sensors, the network sizes were intentionally kept small to reduce computational cost and the likelihood of overfitting. Although the subsequently trained networks were not fully optimized, they yielded reasonable performances and served as a point of comparison across the investigated sensors. For further network refinement, larger hidden layers and stacking LSTM layers could improve detection performance.

\section{Sensor Parameters}
\label{sec:sensor_paramters}
For any tactile sensor, there are typically two controllable parameters of interest: window size and sampling rate. Window size dictates the number of sequential sensor readings that are passed to a classifier before a prediction is made. This parameter is particularly relevant for slip detection since stick-slip events across a tactile sensor are captured over a sequence of readings. Therefore, window sizes must be sufficiently large to capture the slip phenomenon. However, window size also directly inflates slip reaction time. As such, a minimized window size that affords high slip detection accuracy rates is crucial to improving slip reflexive behaviors during grasping or manipulation processes. Sensor sampling rates are often user-selectable and dictate how many sensor readings occur per second. High sampling rates are generally required for quality slip detection. However, minimizing sampling rates while preserving slip detection accuracy is pertinent to minimizing data acquisition resources and will assist tactile sensors with large numbers of sensory signals that cannot be sampled at excessively high sampling rates (e.g., arrays) \cite{alcazar2012estimating,fernandez2014micro,heyneman2016slip}.

\subsection{Window Size}
\label{subsec:window_size}
LSTM networks were trained on the balanced datasets shuffled at six different window sizes -- 5, 10, 25, 50, 100, and 200 -- for data collected at the maximum sensor sampling rate (see Table \ref{tab:window_size}). Overall, the Nano17s and OptoForce20s exhibited the highest slip classification accuracies of over 90 \% after 200 consecutive sensor samples were passed to the LSTM networks. In contrast, the Biotac SPs yielded over 85 \% slip classification rates. The elevated classification performance for the Nano17s and OptoForce20s (when compared to the Biotac SPs) correlated to both the larger disparities in slip state bands as discussed in Section \ref{sec:spectral_analysis} and their finer sensory resolution. Across all sensors, it is unclear whether the classification accuracies obtained from a window size above 50 samples yielded statistically significant improvements. However, accuracy generally decreased with window sizes below 50 samples. With as few as five samples, accuracy dropped by 15 \% for the Nano17s and OptoForce20s and 20 \% for the Biotac SPs. Therefore, to decrease slip prediction time and maintain near-maximum accuracy, a window size of 50 samples yielding a slip detection time of 0.05 sec for the Nano17s and OptoForce20s and 0.06 sec for the Biotac SPs may prove adequate as these times are similar to human slip reflexes \cite{Johansson1984}. To determine the existence of performance bias at this windows size, the true positive (TP, detection of slip during its occurrence) and true negative (TN, detection of non-slip during its occurrence) rates averaged across all fingers were calculated as 87.3 \%, 90.1 \%, and 84.7 \% (TP), and 92.3 \%, 89.79 \%, and 87.0 \% (TN) for the Nano17s, OptoForce20s, and Biotac SPs, respectively. Overall, the TP and TN rates are relatively close across all sensors indicating at most only a marginal bias based on sensor type (an advantage likely incited by data balancing in Section \ref{subsec:data_balancing}).

As a reference point, Holweg's thresholding technique \cite{holweg1996slip}, applied across the Nano17s, OptoForce20s, and Biotac SPs for a window size of 50 samples, yielded classification accuracies of 71.4 \%, 80.9 \%, and 76.4 \%, respectively. Threshold values were determined in two steps: 1) the sum of signal amplitudes (from spectral analysis) over the most significant frequency band (identified in Section \ref{sec:spectral_analysis}) was calculated for every window of data in the non-slipping and slipping datasets; and 2) the location of largest separation in the empirical cumulative distribution functions of these sums from non-slipping and slipping data constituted the threshold value. Threshold values were 0.0015 N, 3.5 units, and 25 units for the Nano17s, OptoForce20s, and Biotac SPs, respectively. Although simpler, this approach yielded classification accuracies that are 10 \% to 20 \% less than that of LSTMs.

\begin{table}[!t]
\caption{Slip state classification accuracy at various window sizes across three types of sensors and fingers.}
\centering
\begin{tabular}{|C{0.85cm} | C{0.85cm} | C{0.6cm} | C{0.6cm} | C{0.6cm} | C{0.6cm}| C{0.6cm}| C{0.6cm}|}
\hline
\multirow{2}{*}{\textbf{Sensor}} & \multirow{2}{*}{\textbf{Finger}} & \multicolumn{6}{c|}{\textbf{Classification Accuracy (\%) at Window Size}}\\
\cline{3-8}
  &  & \bigstrut[t] 5 & \bigstrut[t] 10 & \bigstrut[t] 25 & \bigstrut[t] 50  & \bigstrut[t] 100  & \bigstrut[t] 200 \\ 
\hline 
\multirow{3}{0.7cm}{Nano17} & \bigstrut[t] Index & 74.7 & 80.9 & \bigstrut[t]  85.5 & \bigstrut[t] 88.0 & \bigstrut[t] 89.6 & 90.4 \\
\cline{2-8}
& Middle & 80.3 &  86.2 & 91.0 & 89.9 & 91.5 & 94.2\\
\cline{2-8}
& Little & 84.3 & 86.4 & 91.4 & 91.5 & 92.6 & 92.0\\
\hline
\multirow{3}{*}{\parbox{0.7cm}{Opto- Force20}} & \bigstrut[t] Index & 80.3 & 86.0 & \bigstrut[t]  90.3 & \bigstrut[t]  91.3  & \bigstrut[t] 91.0 & 91.7 \\
\cline{2-8}
& Middle & 82.5 & 87.1 & 91.0 & 91.0 & 92.1 & 91.0\\
\cline{2-8}
& Little & 73.0 & 79.4 & 85.8 & 87.2 & 90.1 & 91.6\\
\hline
\multirow{2}{*}{\parbox{0.7cm}{Biotac SP}} & \bigstrut[t] Index & 66.4 & 74.8 & \bigstrut[t] 82.5 & \bigstrut[t] 86.5 & \bigstrut[t] 86.1 &  86.2\\
\cline{2-8}
& Middle & 68.3 & 73.4 & 80.9 & 85.1 & 88.1 & 85.1 \\
\hline
\end{tabular}
\label{tab:window_size}
\end{table}

\subsection{Sampling Rate}
The LSTM networks were trained with various data sampling rates to measure the impact on classification accuracy. Since the datasets can be sequentially downsampled by a factor of two by omitting every other sensor reading within a sample sequence, five additional sampling rates were investigated as shown in Table \ref{tab:sampling_rate}. The datasets with a window size of 200 samples were chosen for downsampling such that the number of samples per sequence at the lowest sampling rate was greater than five (i.e., a sequence of appreciable length).

Classification accuracy is highly sensitive with respect to sampling rate. For the Nano17s and OptoForce20s, every increase in sampling rate by a factor of two yielded an approximate 2 \% to 4 \% gain in classification accuracy. This consistent and steady ascent in accuracy is explained by Fig. \ref{fig:spectral_analysis}. These sensors had significant disparities in spectral content between non-slipping and slipping data across the full range of available frequencies. Consequently, heightened sampling rates resulted in the continual acquisition of relevant features at higher frequencies in the signal. Notably, the largest gains in accuracy were achieved within a sampling rate of 125 Hz. At this Nyquist frequency, signals with frequencies at approximately 62.5 Hz could be discerned without aliasing effects. Both the Nano17s and OptoForce20s experienced the clearest disparity in signal amplitudes at precisely this frequency. Overall, these two sensors still yielded useful accuracies above 80 \%, even with a factor of ten reduction in maximum sampling rate (i.e., at 125 Hz). This indicated that useful slip detection accuracies were still tenable at relatively low frequencies, an attractive trait for tactile sensors with a large number of sensory outputs.

For the Biotac SPs, significant increases in accuracy were achieved when moving from a sampling rate of 212.5 Hz to 425 Hz. Referring to Fig. \ref{fig:spectral_analysis}, Biotac SP yielded a significant difference in spectral content between 175 Hz and 225 Hz. With a Nyquist frequency of 425 Hz, signals with frequencies up to 212.5 Hz could be seen without aliasing effects. However, at a sampling rate of 212.5 Hz, the significant signal content within 175 Hz and 225 Hz led to the approximate 15 \% loss in classification accuracy. Again, steady increases in classification accuracy of 2 \% to 4 \% were sustained for every increase in sampling rate by a factor of two. This trend is logical, considering the first significant band of frequencies spanned 0 Hz to 75 Hz. In contrast to the Nano17s and OptoForce20s, the Biotac SPs required higher sampling rates of 425 Hz to achieve reasonable slip detection accuracies.

\begin{table}[!t]
\caption{Slip state classification accuracy at various sampling rates across three types of sensors and fingers. *Superscripts indicate a particular sensor type for selected sampling rate: 1 for Nano17s, 2 for OptoForce20s, and 3 for Biotac SPs.}
\centering
\begin{tabular}{|C{0.85cm} | C{0.90cm} | C{0.85cm} | C{0.8cm} | C{0.8cm} | C{0.75cm}| C{0.75cm}|}
\hline
\multirow{2}{*}{\textbf{Sensor}} & \multirow{2}{*}{\textbf{Finger}} & \multicolumn{5}{c|}{\textbf{Classification Accuracy (\%) at Sampling Rate}}\\
\cline{3-7}
  &  & \specialcell{$31.25^{1,2}$ \\ $26.56^3$}& \specialcell{$62.5^{1,2}$ \\ $53.125^3$} & \specialcell{$125^{1,2}$ \\ $106.25^3$}  & \specialcell{$250^{1,2}$ \\ $212.5^3$}  & \specialcell{$500^{1,2}$ \\ $425^3$} \\ 
\hline 
\multirow{3}{*}{Nano17} & \bigstrut[t] Index & 71.6 & \bigstrut[t] 75.2 & \bigstrut[t] 80.0 & \bigstrut[t] 81.6 & 81.6\\
\cline{2-7}
& Middle & 75.8 & 79.7 & 82.2 & 86.6 & 88.6\\
\cline{2-7}
& Little & 80.0 & 85.1 & 89.1 & 87.6 & 89.6\\
\hline
\multirow{3}{*}{\parbox{0.7cm}{Opto- Force20}} & \bigstrut[t] Index & 80.2 & \bigstrut[t] 83.0 & \bigstrut[t] 85.7 & \bigstrut[t] 87.3 &  88.5\\
\cline{2-7}
& Middle & 83.1 & 85.8 & 87.8 & 89.0 & 89.4\\
\cline{2-7}
& Little & 71.4 & 74.7 & 78.2 & 81.0 & 83.8\\
\hline
\multirow{2}{*}{\parbox{0.7cm}{Biotac SP}} & \bigstrut[t] Index & 57.6 & \bigstrut[t] 58.6 & \bigstrut[t] 60.7 & \bigstrut[t] 64.4 &  81.0\\
\cline{2-7}
& Middle & 58.1 & 59.9 & 63.1 & 66.7 & 78.4\\
\hline
\end{tabular}
\label{tab:sampling_rate}
\end{table}

\section{Exogenous Effects}
\label{sec:material_speed_effects}
A truly robust tactile sensor should yield minimally varying slip detection accuracies regardless of material and slip speed effects. However, different materials possess differing tribological properties and slip speed is known to positively correlate with increasing signal amplitude \cite{howe1989sensing}. Therefore, the previously calibrated slip detectors trained with a window size of 50 samples and at the sensors' maximum sampling rate (settings determined to yield maximum classification accuracy with quickest detection time) were tested against these factors.

\subsection{Material}
Both slipping data collected at the four speed settings and non-slipping data for each of the five material types were pooled per material. The previously trained slip detectors were tested on these material-centric datasets to measure material sensitivity. New slip detectors were also trained and tested, but with direct exclusion of data acquired at the evaluation material (see Table \ref{tab:material}). The Nano17s experienced relatively minor fluctuations in slip classification accuracy among all material types (regardless of training scenario). The noticeable exception was neoprene, which yielded multiple accuracy rates below 80 \% with even further losses beyond 20 \% when not explicitly training on neoprene. The Nano17 slip signatures were likely more obscured on this softer material. Similarly, the OptoForce20s were mostly invariant to material type except for cardboard, which had multiple accuracy rates below 80 \% with explicit training and below 60 \% without explicit training. During experimentation, OptoForce20 noticeably gathered cardboard fibers on its rubberized contact surfaces, which likely negatively impacted its tactile sensations for predicting slip. This motivates the need for periodically cleaning tactile sensor surfaces. The Biotac SPs were the most invariant to material type, with small fluctuations in accuracy around a nominal of 80 \% regardless of training scenario. Overall, slip detection accuracy was fairly invariant to material selection for all three sensor types. However, if negatively outlying performance is detected for a particular material, then explicit training on that material and materials of that type will maximize and unify slip detection performance.

\begin{table}[!t]
\caption{Slip state classification accuracy with and without (*) explicit training of indicated materials across sensors.}
\centering
\begin{tabular}{|C{1.4cm} | C{0.85cm} | C{.65cm} | C{0.65cm} | C{0.65cm} | C{0.65cm}| C{0.65cm}|}
\hline
\multirow{2}{*}{\textbf{Sensor}} & \multirow{2}{*}{\textbf{Finger}} & \multicolumn{5}{c|}{\textbf{Classification Accuracy (\%) with Material}}\\
\cline{3-7}
  &  & Aluminum & PVC & Neoprene  & Cardboard  & Plywood \\ 
\hline 
\multirow{3}{*}{Nano17} & \bigstrut[t] Index & 90.9 89.3* & \bigstrut[t] 87.1 84.4* & \bigstrut[t] 76.1 53.7* & \bigstrut[t] 91.1 88.3* & 81.2 76.4*\\
\cline{2-7}
& Middle & 92.8 93.4* & 88.6 87.0* & 79.6 43.4* & 92.2 90.9* & 90.9 89.7*\\
\cline{2-7}
& Little & 90.7 91.3* & 89.3 85.3* & 86.0 80.6* & 90.4 90.0* & 91.4 91.0* \\
\hline
\multirow{3}{*}{OptoForce20} & \bigstrut[t] Index & 86.5 66.8* & \bigstrut[t] 89.5 87.9* & \bigstrut[t] 91.9 90.2*& \bigstrut[t] 77.0 57.2*& 86.9 86.3*\\
\cline{2-7}
& Middle & 93.7 94.5* & 92.7 92.5*& 92.1 88.4*& 67.9 60.5*& 90.0 90.0*\\
\cline{2-7}
& Little & 90.5 84.8* & 86.0 85.4*& 88.9 86.5*& 86.6 86.2* & 80.6 82.2*\\
\hline
\multirow{2}{*}{Biotac SP} & \bigstrut[t] Index & 81.1 78.9*& \bigstrut[t] 83.2  82.4*& \bigstrut[t] 84.6 83.8*& \bigstrut[t] 79.3 76.4*& 79.6 78.5*\\
\cline{2-7}
& Middle & 81.1 80.6*& 80.8 80.2*& 80.9 80.4*& 79.4 77.0*& 81.1 80.8*\\
\hline
\end{tabular}
\label{tab:material}
\end{table}

\subsection{Slip Speed}
Both non-slipping and slipping data collected across the five materials were pooled for each of the four arm speed settings. The previously trained slip detectors were tested on these speed-centric datasets to measure their performance sensitivity to slip speed effects. Furthermore, new slip detectors were trained on the pooled dataset with explicit exclusion of data acquired at the evaluation speed. Table \ref{tab:speed} reports the classification accuracies obtained with and without explicit training on the evaluation speeds. The greatest loss in slip classification accuracy across all sensor types was experienced at the smallest slip speeds (5 mm/s). Without explicit training on data acquired at this speed, both the OptoForce20s and Biotac SPs persistently experienced a further reduction of classification accuracy. Conclusively, slip signal signal strength is diminished at 5 mm/s, yielding reduced accuracies. To maximize performance, explicit training on slow slip speeds should be conducted. In contrast, classification accuracies remained relatively consistent at all other speeds regardless of their explicit training. Overall, the Nano17s produced the most resilient slip detectors across slip speeds, with almost all classification accuracies above 80 \%. OptoForce20s behaved similarly, but with many accuracies below 80 \% at 5 mm/s. Biotac SPs were the most sensitive to slip speed with several classification accuracies below 70 \% at 5 mm/s.

\begin{table}[!t]
\caption{Slip state classification accuracy with and without (*) explicit training of indicated speeds across sensors.}
\centering
\begin{tabular}{|C{1.4cm} | C{0.90cm} | C{.9cm} | C{1cm} | C{1cm} | C{1cm}|}
\hline
\multirow{2}{*}{\textbf{Sensor}} & \multirow{2}{*}{\textbf{Finger}} & \multicolumn{4}{c|}{\textbf{Classification Accuracy (\%) at Speed}}\\
\cline{3-6}
  &  & 5 mm/s & 25 mm/s  & 50 mm/s  & 75 mm/s \\ 
\hline 
\multirow{3}{*}{Nano17} & \bigstrut[t] Index & 80.8 70.4*& \bigstrut[t] 86.7 86.5*& \bigstrut[t] 86.6 86.7*& \bigstrut[t] 87.1 84.8*\\
\cline{2-6}
& Middle & 82.6 85.8* & 89.8 92.4*& 91.0 90.9*& 91.6 92.5*\\
\cline{2-6}
& Little & 82.5 83.7*& 90.6 90.3*& 91.7 90.8* & 92.5 91.8*\\
\hline
\multirow{3}{*}{OptoForce20} & \bigstrut[t] Index & 80.1 76.7*& \bigstrut[t] 87.6 87.8*& \bigstrut[t] 88.8 89.1*& \bigstrut[t] 88.9 89.0*\\
\cline{2-6}
& Middle & 82.6 77.0*& 87.8 90.9*& 88.8 88.8*& 89.2 89.1*\\
\cline{2-6}
& Little & 75.3 73.9*& 88.6 89.7*& 91.1 91.1*& 91.5 91.7*\\
\hline
\multirow{2}{*}{Biotac SP} & \bigstrut[t] Index & 71.4 62.4*& \bigstrut[t] 84.1 84.5*& \bigstrut[t] 85.5 84.7*& \bigstrut[t] 85.2 82.2*\\
\cline{2-6}
& Middle & 67.6 59.6*& 83.8 84.1*& 86.1 84.9*& 86.1 84.5*\\
\hline
\end{tabular}
\label{tab:speed}
\end{table}

\section{Manufacturing Variability}
\label{sec:manu_Var}

The effect of manufacturing variability directly impacts the cost of calibrating a sensor for robust slip detection. Calibration costs due to data collection and sensor repair are drastically reduced if the calibration models of one sensor can be directly re-applied to another sensor of the same type without any additional data collection. To quantify the impact of manufacturing variability and model generality, a slip detection model trained using data gathered from a sensor on one finger was applied to data of the same sensor type on another finger. The models used were again those previously trained with a window size of 50 samples and at the sensors' maximum sampling rate. Additionally, a new model was trained per sensor type on data acquired from all instances of that sensor type (a ``combined'' dataset). Evaluating this model on both the individual and combined datasets provided insight on performance gains from batching sensory data acquired across multiple instances of the same sensor type.” Table \ref{tab:man_var_all} shows the results for all sensor types.

Encouragingly, all sensor types exhibited a relatively high level of model transference. Loss in classification accuracy remained below 11 \% for the Nano17s, 21 \% for the OptoForce20s, and 3 \% for the Biotac SPs. This result appears to indicate that the Biotac SPs had the highest level of model transference, followed by the Nano17s and OptoForce20s. However, confidence in this assessment concerning the Biotac SPs is diminished since only two sensors of this type were available. Regardless, training on data collected by all instances of a particular sensor type facilitated the equalization of the classification accuracy. The Nano17s, OptoForce20s, and Biotac SPs exhibited a classification difference within 5 \%, 3 \%, and 2 \%, respectively, among all sensor instances. Conclusively, immediate model transfer among sensors within a sensor type is possible. However, classification accuracy can be further equalized by training with data collected by a batch of sensors. This action will help reduce performance bias towards a particular sensor, and will likely generalize the model more accurately to all future instances of sensors for that particular type. 

\section{Discussion}
Spectral analysis of univariate tactile signal gradients revealed a positive, yet complex disparity in sensory response between non-slipping and slipping stimuli for three sensor types with milli-Newton level sensitivity. Statistically significant spectral content for large frequency bands between non-slipping and slipping signals indicated that the datasets captured relevant vibratory features towards slip detection, independent of absolute shear or contact force magnitudes, and contact location. This observation corroborates neurophysiological findings that contact phenomena like slipping produces high-frequency mechanical vibrations through tactile surfaces, requiring highly-sensitive receptors for proper detection. These particular mechanoreceptors are also insensitive to contact forces and locations.

When trained on the these univariate tactile signals, a simple thresholding method yielded slip classification accuracies around 70 \% - 80 \%, while those of LSTM networks produced accuracies above 90 \% for most sensors. This result indicated that the underlying slip phenomena's encodings were very complex and required sophisticated machine learning models to produce more robust slip classification rates. Overall, the performance measurement methodology uncovered that larger window sizes and higher sampling frequencies improved slip detection accuracies. The trained LSTM models also exhibited robustness to slip speed, material types, and manufacturing variability with the exception of a few edge cases. Since the LSTM models were at most only passed 0.2 sec of sensor gradient signals and the model internal states were reset before each transmission, the magnitude of shear force (or resultant force) is \textit{not} observable by the model. This further confirmed that univariate tactile signal gradients alone can principally encode slip-state mechanics, yielding LSTM models with over 90 \% classification accuracy. Tactile sensors with only a single sensing element can then be calibrated for robust slip detection, motivating simpler and affordable sensor designs.

Arguably, accuracy could still be improved by analyzing groups of tactile signals as reported by Heyneman in \cite{heyneman2016slip}. However, this approach would require spatially distributed tactile outputs sampled at high frequencies, more complex calibration models, and data collection processes (to thoroughly explore multi-output sensor responses). As shown, this approach is not necessary to achieve robust classification accuracy. In fact, LSTMs produced at least 85 \% accuracy rates with the Biotac SPs, an immediate improvement over the 80 \% accuracy rates reported by Heyneman when grouping signals from a single Biotac. Heyneman also reported 99 \% accuracy with a PVDF sensor, but this level of accuracy was only obtained with this type of sensor. Conclusively, the leap in performance was a greater testament to the sensor design itself, and not the algorithmic approach. With the application of LSTMs or similarly powerful machine learning models, multivariate signals can at best only offer marginal improvements to accuracy rates. 

With robust slip detection achieved, future work includes investigating the significance of this sensing modality towards actual grasping and manipulation control policies. Specifically, injecting the trained slip detectors into grasp reflexes or manipulation controllers could improve operational robustness and overall real-world performance, a functional relation already observed in biological systems.

\begin{table}[!htb]
\caption{Classification accuracy with LSTMs trained and tested on data from one or all fingers of a single sensor type.}
\centering
\begin{tabular}{|C{0.2cm} | C{1.5cm} | C{0.8cm} | C{1cm} | C{0.8cm} | C{1.2cm}|N}
\hline
\multicolumn{2}{|c|}{\multirow{2}{*}{\textbf{Training Data}}} & \multicolumn{4}{c|}{\textbf{Classification Accuracy (\%)}}\\
\cline{3-6}
\multicolumn{2}{|c|}{} & Index & Middle & Little & Combined &\\ [1ex]
\hline
\multicolumn{2}{|c|}{Nano17, Index} & 88.0 & 89.5 & 84.7 & 87.5&\\
\multicolumn{2}{|c|}{Nano17, Middle} & 82.5 & 89.9 & 86.1 & 86.2&\\
\multicolumn{2}{|c|}{Nano17, Little} & 80.5 & 83.7 & 91.5 & 85.2&\\
\multicolumn{2}{|c|}{Nano17, Combined} & 84.8 & 89.1 & 89.1 & 87.8&\\
\hline
\multicolumn{2}{|c|}{OptoForce20, Index} & 91.3 & 89.9 & 73.0 & 84.8&\\
\multicolumn{2}{|c|}{OptoForce20, Middle} & 87.7 & 91.0 & 70.4 & 83.2&\\
\multicolumn{2}{|c|}{OptoForce20, Little} & 85.3 & 87.6 & 87.2 & 86.7&\\
\multicolumn{2}{|c|}{OptoForce20, Combined} & 91.5 & 92.2 & 89.4 & 91.0&\\
\hline
\multicolumn{2}{|c|}{Biotac SP, Index} & 86.5 & 83.5 & NA & 85.1&\\
\multicolumn{2}{|c|}{Biotac SP, Middle} & 86.0 & 85.1 & NA & 85.7&\\
\multicolumn{2}{|c|}{Biotac SP, Combined} & 86.9 & 85.4 & NA & 86.3&\\
\hline

\end{tabular}
\label{tab:man_var_all}
\end{table}

\bibliographystyle{IEEEtran}
\bibliography{References}

\end{document}